%% file: main.tex
\documentclass{aip-book}

\usepackage{natbib}
\usepackage{sec/package}

\usepackage{graphicx,hologo,mwe,url} % for elements of the example chapter
\usepackage{array,booktabs,threeparttable} % useful for the table example
\usepackage{amsmath}

\usepackage{amsthm}

\theoremstyle{definition}

\usepackage{fancyhdr}
\fancypagestyle{alim}{\fancyhf{}\fancyfoot[L]{
\textit{Handbook of Computational Social Science for Policy. 2023. Chapter 7, pages 141 -- 162.
    \\
    Open Access on Springer: \url{https://doi.org/10.1007/978-3-031-16624-2}
    }
    \\
    \hfill 1 \hfill
    }
    }

\begin{document}
\setchapter{7} % set this to your chapter number (this will be provided during final typesetting)
\chapter{Natural Language Processing for Policymaking} % the title of your chapter

% each author's information is entered with the \chapterauthor macro
% this section is only needed for "contributed" books
% for "authored" books, comment out or remove the \chapterauthor commands
\chapterauthor{Zhijing Jin}{jinzhi@ethz.ch}{Max Planck Institute \& ETH Zürich}

\chapterauthor{Rada Mihalcea}{mihalcea@umich.edu}{University of Michigan}

\begin{abstract}
Language is the medium for many political activities, from campaigns to news reports. Natural language processing (NLP) uses computational tools to parse text into key information that is needed for policymaking. In this chapter, we introduce common methods of NLP, including text classification, topic modeling, event extraction, and text scaling. We then overview how these methods can be used for policymaking through four major applications including data collection for evidence-based policymaking, interpretation of political decisions, policy communication, and investigation of policy effects.
Finally, we highlight some potential limitations and ethical concerns when using NLP for policymaking.
% Finally, we will highlight the challenges of NLP for policymaking in the coming era.

\end{abstract}


\thispagestyle{alim}

\input{main_content}
\bibliographystyle{acl_natbib}
\bibliography{main}

\end{document}

%% file: main_content.tex
\section{Introduction}\label{sim:sec:intro}

\begin{figure}[t]
    \centering
    \includegraphics[width=0.8\textwidth]{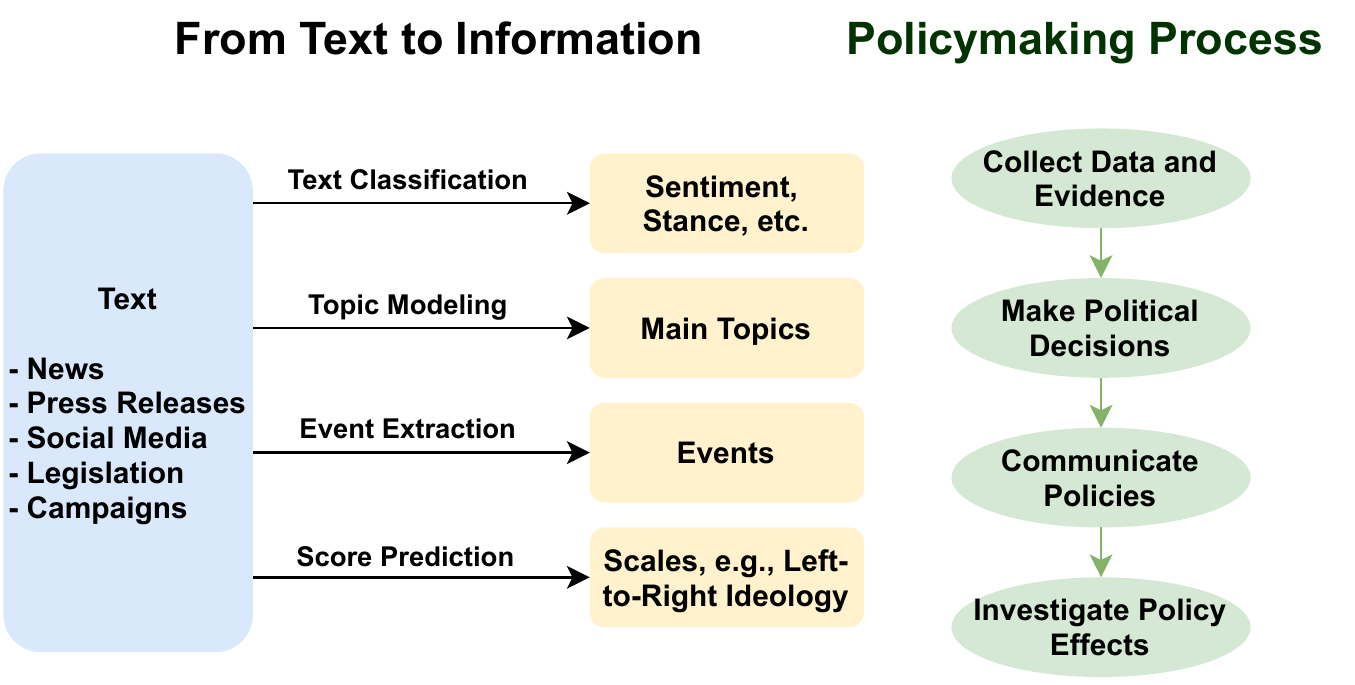}
    \caption{Overview of NLP for policymaking.}
    \label{fig:intro}
\end{figure}

Language is an important form of data in politics.  Constituents express their stances and needs in text such as social media and survey responses. Politicians conduct campaigns through debates, statements of policy positions, and social media. Government staff needs to compile information from various documents to assist in decision-making. Textual data is also prevalent through the documents and debates in the legislation process, negotiations and treaties to resolve international conflicts, and media such as news reports, social media, party platforms, and manifestos.

% Language is the medium for politics and political conflict. Candidates debate and state policy
% positions during a campaign. Once elected, representatives write and debate legislation. After
% laws are passed, bureaucrats solicit comments before they issue regulations. Nations regularly
% negotiate and then sign peace treaties, with language that signals the motivations and relative
% power of the countries involved. News reports document the day-to-day affairs of international
% relations that provide a detailed picture of conflict and cooperation. Individual candidates and
% political parties articulate their views through party platforms and manifestos. Terrorist groups
% even reveal their preferences and goals through recruiting materials, magazines, and public statements.
Natural language processing (NLP) is the study of computational methods to automatically analyze text and extract meaningful information for subsequent analysis.
The importance of NLP for policymaking has been highlighted since the last century \citep{gigley-1993-projected}. With the recent success of NLP and its versatility over tasks such as classification, information extraction, summarization, and translation \citep{devlin2019bert,brown2020language}, there is a rising trend to integrate NLP into the policy decisions and public administrations \citep{misuraca2020use,engstrom2020government,vanroy2021ai}. Main applications include extracting useful, condensed information from free-form text \citep{engstrom2020government}, and 
% For example, Germany seeks to integrate chatbots into the digital infrastructure of customs administration.
% the AI Watch report by European Commissions \citep{vanroy2021ai} 
analyzing sentiment and citizen feedback by NLP \citep{biran2022policycloud} as in many projects funded by EU Horizon projects \citep{european2011proposal}.
Driven by the broad applications of NLP \citep{jin-etal-2021-good,gonzalez2022good}, the research community also starts to connect NLP with various social applications in
% there is an increasing amount of attention on adopting NLP methods to
the fields of computational social science \citep{lazer2009computational,shah2015big,engel2021handbook,luz2022computational} and political science in particular \citep{grimmer2013text,glavas-etal-2019-computational}.
% \footnote{Apart from the increasing number of papers published on NLP for political science, there are also activities such as the tutorial on NLP and political science \citep{glavas-etal-2019-computational} and hackathon \citep{nanni2018findings}.}

% The importance of NLP for government is already highlighted \citep{gigley-1993-projected} for communication needs and knowledge compilation.

% The recent adoption of NLP methods had led to
% significant advances in the field of Computational
% Social Science (CSS) \citep{lazer2009life} and political science in particular \citep{grimmer2013text}.

We show an overview of NLP for policymaking in Figure~\ref{fig:intro}. According to this overview, the chapter will consist of three parts. First, we introduce in Section~\ref{sec:nlp_tools} NLP methods that are applicable to political science, including text classification, topic modeling, event extraction, and score prediction. Next, we cover a variety of cases where NLP can be applied to policymaking in Section~\ref{sec:nlp4policy}. Specifically, we cover four stages: analyzing data for evidence-based policymaking, improving policy communication with the public, investigating policy effects, and interpreting political phenomena to the public. Finally, we will discuss limitations and ethical considerations when using NLP for policymaking in Section~\ref{sec:ethics}.
% In the end, we will also highlight several important challenges of NLP for policymaking, including the caveat of misinformation, disinformation, and propaganda, cybersecurity problem for important policy decisions, inter-group conflicts and how to promote mutual understanding, and international competitions in Section~\ref{sec:challenges}.

% \ea\label{sim:ex:speaker}
% \ea \sib{The speaker is drunk}$^{g,c}=\textsc{drunk}\big(\iota x\,\textsc{speaker}(x)\big)$\label{sim:ex:speaker-a}
% \ex \sib{I am drunk}$^{g,c}=\textsc{drunk}\big(\cnst{speaker}(c)\big)$\label{sim:ex:speaker-b}
% \z\z

\section{NLP for Text Analysis}\label{sec:nlp_tools}

NLP brings powerful computational tools to analyze textual data \citep{jurafsky2000speech}.
According to the type of information that we want to extract from the text, we introduce four different NLP tools to analyze text data: text classification (by which the extracted information is the \textit{category} of the text), topic modeling (by which the extracted information is the \textit{key topics} in the text), event extraction (by which the extracted information is the list of \textit{events} mentioned in the text), and score prediction (where the extracted information is a \textit{score} of the text). Table~\ref{tab:nlp_tasks} lists each method with the type of information it can extract and some example application scenarios, which we will detail in the following subsections.

\begin{table}[t]
    \centering
    \small
\resizebox{\textwidth}{!}{
    \begin{tabular}{llll}
\toprule
\textbf{NLP Method} & \textbf{Information to Extract} & \textbf{Example Applications} \\ \midrule
Text classification & Category of text & Identify the sentiment, stance, etc.
\\
Topic modeling & Key topics in text & Summarize topics in political agenda
\\
Event extraction & List of events & Extract news events, international conflicts
\\
Score prediction & Score & Text scaling
\\
\bottomrule
    \end{tabular}
}
    \caption{Four common NLP methods, the type of information extracted by each of them, and example applications.}
    \label{tab:nlp_tasks}
\end{table}

\subsection{Text Classification}
\begin{figure}
    \centering
    \includegraphics[width=0.9 \textwidth]{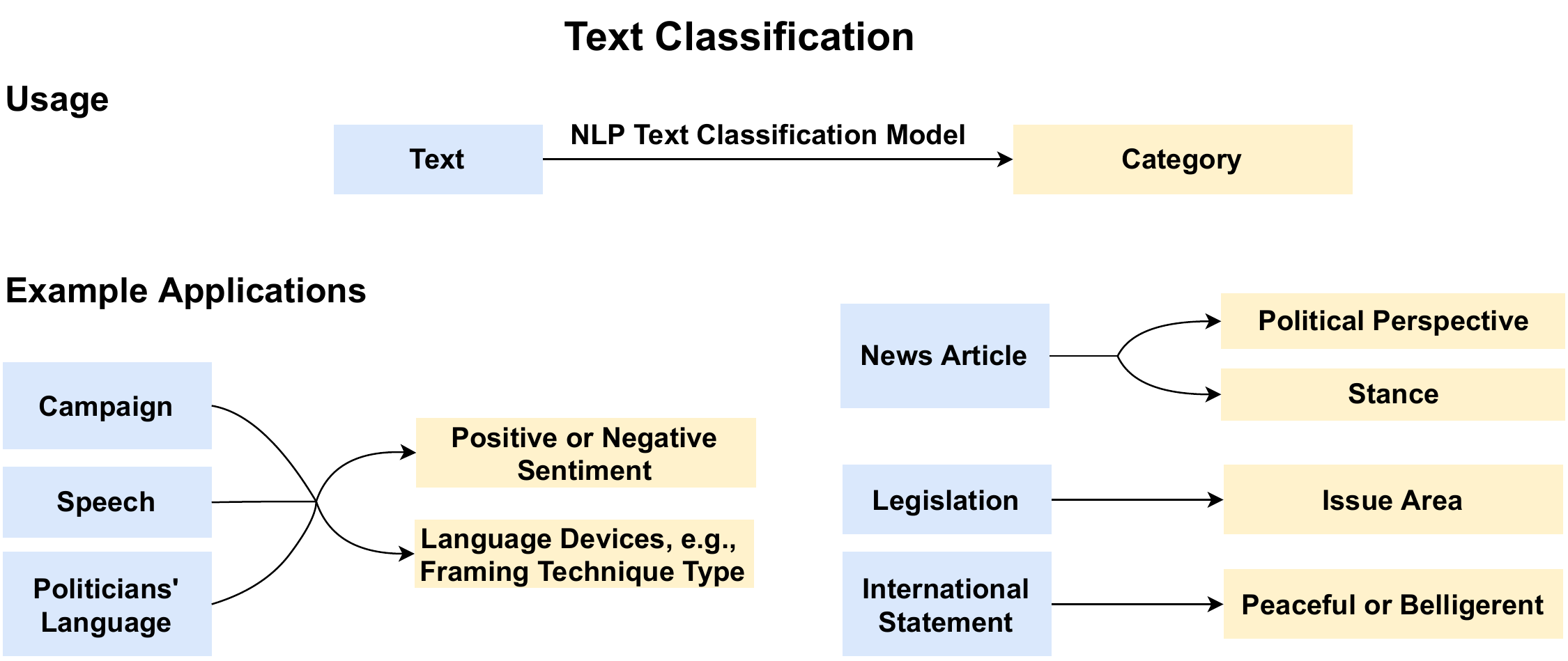}
    \caption{The usage and example applications of text classification on political text.}
    \label{fig:classification}
\end{figure}

As one of the most common types of text analysis methods, text classification reads in a piece of text and predicts its category using an NLP text classification model, as in Figure~\ref{fig:classification}.

There are many off-the-shelf existing tools for text classification \citep{yin-etal-2019-benchmarking,brown2020language,loria2018textblob} such as {the implementation}\footnote{\url{https://discuss.huggingface.co/t/new-pipeline-for-zero-shot-text-classification/681}} using the Python package  \texttt{transformers} \citep{wolf-etal-2020-transformers}. 
A well-known subtask of text classification is sentiment classification (also known as sentiment analysis, or opinion mining), which aims to distinguish the subjective information in the text, such as positive or negative sentiment \citep{pang2007opinion}.
However, the existing tools only do well in categories that are easy to predict. If the categorization is customized and very specific to a study context, then there are two common solutions. One is to use dictionary-based methods, by a list of frequent keywords that correspond to a certain category \citep{albaugh2013automated} or using general linguistic dictionaries such as the Linguistic Inquiry and Word Count (LIWC) dictionary \citep{pennebaker2001linguistic}.
The second way is to adopt the data-driven pipeline, which requires human hand coding of documents into a predetermined set of categories, then train an NLP model to learn the text classification task \citep{sun2019finetune}, and verify the performance of the NLP model on a held-out subset of the data, as introduced in \citet{grimmer2013text}. An example of adapting the state-of-the-art NLP models on a customized dataset is demonstrated in {this guide}.\footnote{\url{https://skimai.com/fine-tuning-bert-for-sentiment-analysis/}}

Using the text classification method, we can automate many types of analyses in political science. As listed in the examples in Figure~\ref{fig:classification}, researchers can detect political perspective of news articles \citep{huguet-cabot-etal-2020-pragmatics}, the stance in media on a certain topic \citep{luo-etal-2020-detecting}, whether campaigns use positive or negative sentiment \citep{ansolabehere1995going}, which issue area is the legislation about \citep{adler2011congressional},
topics in parliament speech \citep{albaugh2013automated,osnabrugge2021cross}, congressional bills \citep{hillard2008computer,collingwood2012tradeoffs} and political agenda \citep{karan-etal-2016-analysis},
whether the international statement is peaceful or belligerent \citep{schrodt2000pattern}, whether a speech contains positive or negative sentiment \citep{schumacher2016euspeech}, and whether a U.S. Circuit Courts case decision is conservative or liberal \citep{hausladen2020text}.
Moreover, text classification can also be used to categorize the type of language devices that politicians use, such as what type of framing the text uses \citep{huguet-cabot-etal-2020-pragmatics}, and 
whether a tweet uses political parody \citep{maronikolakis-etal-2020-analyzing}.

% For example, researchers may ask if campaigns issue positive or negative advertisements
% \citep{ansolabehere1995going}, if legislation is about the environment or some other issue area \citep{adler2011congressional}, if international statements are belligerent or peaceful \citep{schrodt2000pattern},
% or if local news coverage is positive or negative \citep{eshbaugh2010tone}.

% Model political communication strategy, metaphor, emotion and political rhetoric, and demonstrate that they advance performance in three tasks: predicting political perspective of news articles, party affiliation of politicians and framing of policy issues \citep{huguet-cabot-etal-2020-pragmatics}

% Dictionary methods, for example, use the frequency of key words
% to determine a document’s class

% supervised methods.  human hand coding of documents into
% a predetermined set of categories.

% mixed membership models, include problem-specific structure to assist in the estimation
% of categories

% , Computer Assisted Clustering (CAC), provides a
% method to explore thousands of potential clusterings

\subsection{Topic Modeling}
Topic modeling is a method to uncover a list of frequent topics in a corpus of text. For example, news articles that are against vaccination might frequently mention the topic ``autism,'' whereas news articles supporting vaccination will be more likely to mention ``immune'' and ``protective.''
One of the most widely used models is the Latent Dirichlet Allocation (LDA) \citep{blei2001latent} which is available in the Python packages \texttt{NLTK} and \texttt{Gensim}, as in {this guide}.\footnote{\url{https://skimai.com/fine-tuning-bert-for-sentiment-analysis/}} 
\begin{figure}
    \centering
    \includegraphics[width=0.8\textwidth]{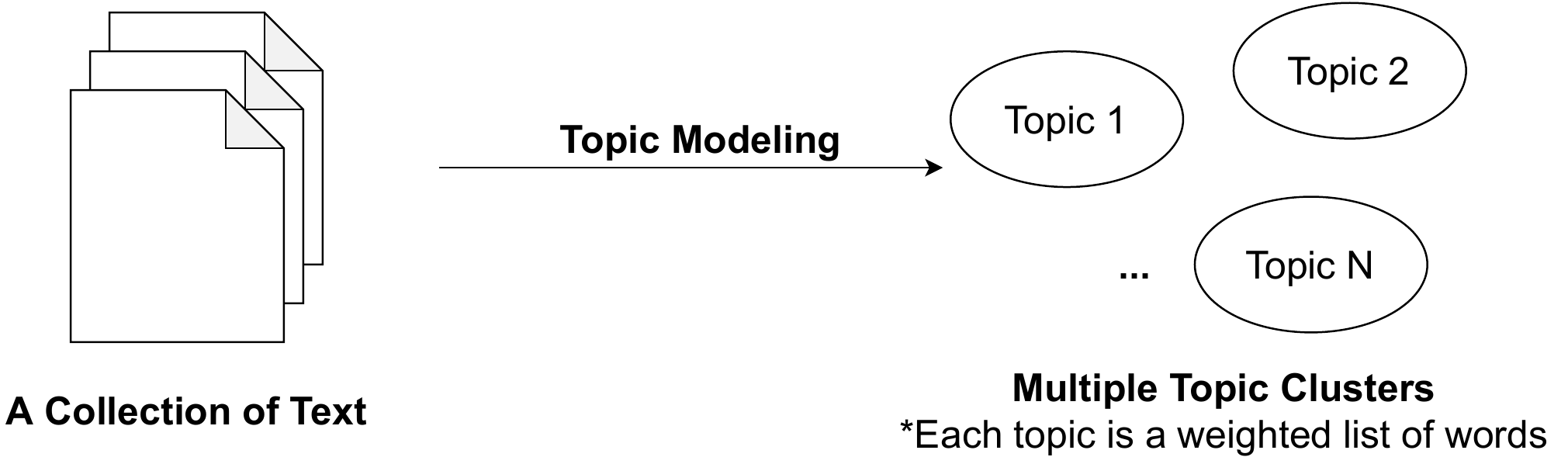}
    \caption{Given a collection of text documents, topic modeling generates a list of topic clusters.}
    \label{fig:topic_model}
\end{figure}

Specifically, LDA is a probabilistic model that models each topic as a mixture of words, and each textual document can be represented as a mixture of topics. As in Figure~\ref{fig:topic_model}, given a collection of textual documents, LDA topic modeling generates a list of topic clusters, for which the number $N$ of topics can be customized by the analyst.  In addition, if needed, LDA can also produce a representation of each document as a weighted list of topics. While often the number of topics is predetermined by the analyst, this number can also be dynamically determined by measuring the perplexity of the resulting topics. In addition to LDA, other topic modeling algorithms have been used extensively, such as those based on principal component analysis (PCA) \citep{Chung2008-fo}.
% Specifically, LDA is a generative probabilistic model that adopts a three-level hierarchical Bayesian model, in which each
% item of a collection is modeled as a finite mixture over an underlying set of topics. Each topic is, in turn, modeled as an infinite mixture over an underlying set of topic probabilities. In the context of
% text modeling, the topic probabilities provide an explicit representation of a document. We present
% efficient approximate inference techniques based on variational methods and an EM algorithm for
% empirical Bayes parameter estimation.

Topic modeling, as described in this section, can facilitate various studies on political text.
% we focus on a large body of work of topical analysis of political texts, covering unsupervised topic induction, including 
Previous studies analyzed the topics of legislative speech \citep{quinn2010analyze,quinn2006automated}, Senate press releases \citep{grimmer2010bayesian}, and electorial manifestos \citep{menini-etal-2017-topic}.

% use unsupervised topic induction such as
% dictionary-based \citep{albaugh2013automated}, topic-modelling and text segmentation \citep{glavas-etal-2016-unsupervised} approaches  \citep{quinn2006automated,quinn2010analyze,grimmer2010bayesian,glavas-etal-2016-unsupervised,menini-etal-2017-topic}
% , as well as supervised topic classification \citep{hillard2008computer,collingwood2012tradeoffs,karan-etal-2016-analysis}. 
% We will also cover more recent work
% on cross-lingual topic classification in political
% texts \citep{glavas-etal-2017-cross,subramanian-etal-2018-hierarchical}. We will further emphasize topic classification models that exploit large manually anotated corpora from CMP \citep{zirn2016classifying,subramanian-etal-2017-joint} and CAP \citep{karan-etal-2016-analysis,albaugh2013automated} projects.

\subsection{Event Extraction}
Event extraction is the task of extracting a list of events from a given text. It
is a subtask of a larger domain of NLP called information extraction \citep{manning2008introduction}.
For example, the sentence ``Israel bombs Hamas sites in Gaza'' expresses an event ``\textit{Israel $\xrightarrow[]{\text{bombs}}$ Hamas sites}'' with the location ``\textit{Gaza}.'' Event extraction usually incorporates both entity extraction (e.g., Israel, Hamas sites, and Gaza in the previous example) and relation extraction (e.g., ``bombs'' in the previous example).

% the NLP community has mainly focused on the extraction of fine-grained events, which constitute n-ary relations
% between entities, such as time and location. For example, an event
% extracted from the sentence “Mr Miller went to Boston in August”
% connects the entities Mr Miller, Boston and August with the predicate “went to”. 

Event extraction is a handy tool to monitor events automatically, such as detecting news events \citep{walker2006ace,mitamura2017events}, 
% international relations \citep{hudson1991artificial}, 
and detecting international conflicts \citep{azar1980conflict,trappl2006programming}. To foster research on event extraction, there are tremendous efforts into textual data collection \citep{mcclelland1976world,schrodt2006twenty,merritt1993international,raleigh_introducing_2010,sundberg2013introducing}, event coding schemes to accommodate different political events \citep{goldstein1992conflict,bond1997mapping,gerner2002conflict}, and dataset validity assessment \citep{schrodt1994validity}.

As for event extraction models, similar to text classification models, there are off-the-shelf tools such as the Python packages \texttt{stanza} \citep{qi2020stanza} and \texttt{spaCy} \citep{honnibal2020spacy}. In case of customized sets of event types, researchers can also train NLP models on a collection of textual documents with event annotations \citep[\textit{inter alia}]{hogenboom2011overview,liu2020extracting}.

\subsection{Score Prediction}
NLP can also be used to predict a score given input text. A useful application is political text scaling, which aims to predict a score (e.g., left-to-right ideology, emotionality, and different attitudes towards the European integration process) for a given piece of text (e.g., political speeches, party manifestos, and social media posts) \citep[\textit{inter alia}]{laver2003extracting,lowe2011scaling,slapin2008scaling,gennaro2021emotion}.

Traditional models for text scaling include Wordscores \citep{laver2003extracting} and WordFish \citep{slapin2008scaling,lowe2011scaling}. Recent NLP models represent the text by high-dimensional vectors learned by neural networks to predict the scores \citep{glavas-etal-2017-unsupervised,nanni2019political}. One way to use the NLP models is to apply off-the-shelf general-purpose models such as InstructGPT \citep{ouyang2022instructGPT} and design a prompt to specify the type of the scaling to the API,\footnote{\url{https://beta.openai.com/docs/introduction}}, or borrow existing, trained NLP models if the same type of scaling has been studied by previous researchers. Another way is to collect a dataset of text with hand-coded scales, and train NLP models to learn to predict the scale, similar to the practice in \citet{slapin2008scaling,gennaro2021emotion}, \textit{inter alia}.

% text scaling task, we will
% present in detail the traditional scaling models
% that operate on lexical text representations such
% as Wordscores \citep{laver2003extracting} and WordFish \citep{slapin2008scaling,lowe2011scaling}
% as well as a more recent scaling approach
% that exploits latent semantic text representations \citep{glavas-etal-2017-unsupervised,nanni2019political}

% \subsection{Representing Text as Vectors: Pretrained Language Models}
% \subsection{Advanced NLP+X Tools}
% \subsubsection{NLP with Causal Inference}
% \citep{egami2018make,feder2021causal}

% \citep{pryzant-etal-2021-causal}
% \subsubsection{NLP with Social Networks}

\section{Using NLP for Policymaking}\label{sec:nlp4policy}

In the political domain, there are large amounts of textual data to analyze \citep{neuendorf2015content}, such as parliament debates \citep{van2017debates}, speeches \citep{schumacher2016euspeech}, legislative text \citep{baumgartner2006comparative,bevan2017gone}, database of political parties worldwide \citep{doring2019party}, and expert survey data \citep{bakker2015measuring}. Since it is tedious to hand-code all textual data, NLP provides a low-cost tool to automatically analyze such massive text. 

In this section, we will introduce how NLP can facilitate four major areas to help policymaking: before policies are made, researchers can use NLP to analyze data and extract key information for evidence-based policymaking (Section~\ref{sec:use1}); after policies are made, researchers can interpret the priorities among and reasons behind political decisions (Section~\ref{sec:use2}); researchers can also analyze features in the language of politicians when communicating the policies to the public (Section~\ref{sec:use3}); finally, after the policies have taken effect, researchers can investigate the effectiveness of the policies (Section~\ref{sec:use4}).

% Data is the prerequisite for many political analyses \citep{schrodt2006twenty}.
% Existing data: 

% International annotation projects: the topically coded electoral programmes (i.e., the Manifesto Corpus) \citep{merz2016manifesto} developed
% within the scope of the Comparative Manifesto
% Project \citep{werner2011manifesto,mikhaylov2012coder} and the topically coded legislative texts annotated for numerous countries within the scope of the Comparative Agenda Project \citep{baumgartner2006comparative,bevan2017gone}.

% European Parliament debates \citep{van2017debates}, EU elite speeches \citep{schumacher2016euspeech}, database of political parties worldwide \citep{doring2019party}, expert survey data \citep{bakker2015measuring},
% other large collection of political texts \citep{bakker2015measuring}

% the inherent difficulties in collecting political texts and political data in general and analyze crowdsourcing as an efficient and agile method for producing political data \citep{benoit2016crowd}.

\subsection{Analyzing Data for Evidence-Based Policymaking}\label{sec:use1}

A major use of NLP is to extract information from large collections of text. This function can be very useful for analyzing the views and needs of constituents, so that policymakers can make decisions accordingly.

\begin{figure}
    \centering
    \includegraphics[width=0.9\textwidth]{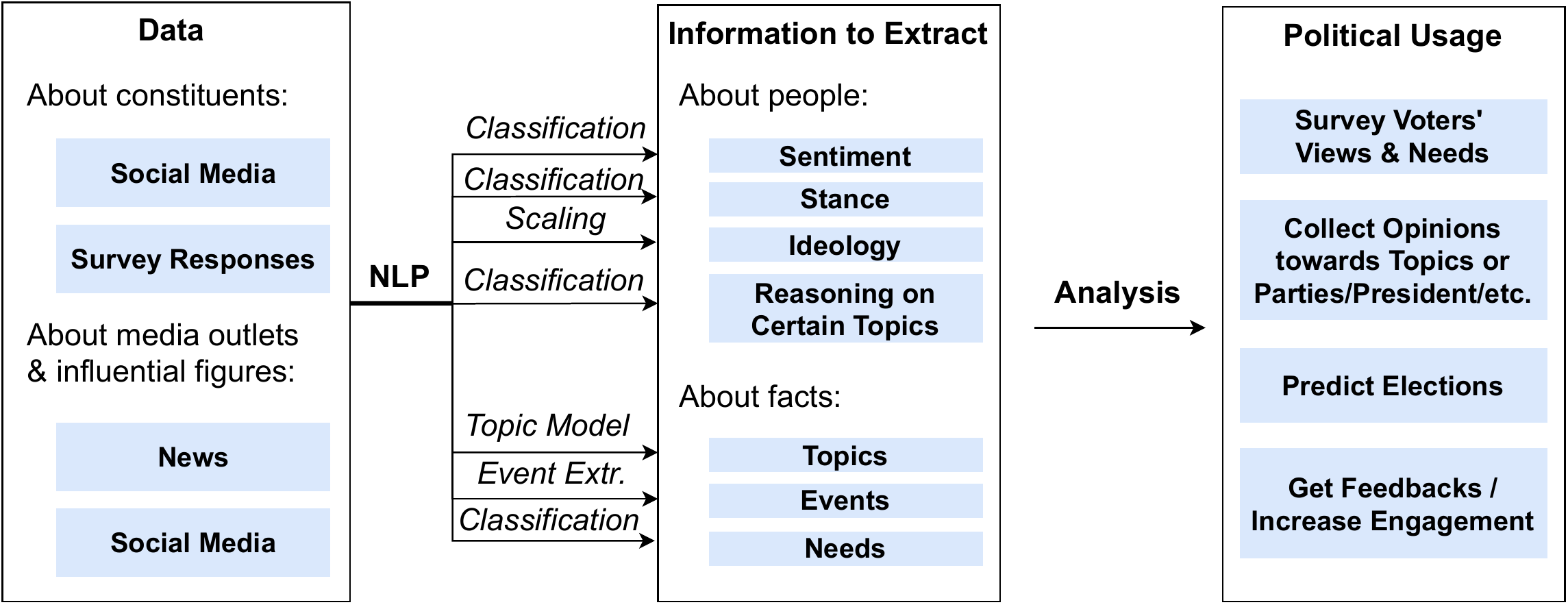}
    \caption{NLP to analyze data for evidence-based policymaking.}
    \label{fig:use_survey}
\end{figure}

As in Figure~\ref{fig:use_survey}, we will explain how NLP can be used to analyze data for evidence-based policymaking from three aspects: data, information to extract, and political usage.

\myparagraph{Data.}
Data is the basis of such analyses. Large amounts of textual data can reveal information about constituents, media outlets, and influential figures. The data can come from a variety of sources, including social media such as Twitter and Facebook, survey responses, and news articles.

\myparagraph{Information to Extract.}
Based on the large textual corpora, NLP models can be used to extract information that are useful for political decision-making, ranging from information about people, such as sentiment \citep{thelwall2011sentiment,rosenthal2015semeval}, stance \citep{thomas-etal-2006-get,gottipati-etal-2013-learning,stefanov-etal-2020-predicting,luo-etal-2020-detecting}, ideology \citep{hirst2010party,iyyer-etal-2014-political,preotiuc-pietro-etal-2017-beyond}, and reasoning on certain topics \citep{egami2018make,demszky-etal-2019-analyzing,camp2021thin}, to factual information, such as main topics \citep{gottipati-etal-2013-learning}, events \citep{trappl2006programming,mitamura2017events,ding-riloff-2018-human,ding-etal-2019-improving}, and needs \citep{sarol-etal-2020-empirical,crayton2020narratives,paul-frank-2019-ranking} expressed in the data. 
The extracted information cannot only be about people, but also about political entities, such as the left-right political scales of parties and political actors \citep{slapin2008scaling,glavas-etal-2017-unsupervised}, which claims are raised by which politicians \citep{blessing-etal-2019-environment,pado-etal-2019-sides}, and the legislative body’s vote breakdown for state bills by backgrounds such as gender, rural-urban and ideological splits \citet{davoodi-etal-2020-understanding}.
% develops a tool realizes the complete workflow necessary for annotating a large newspaper text collection with rich information about claims (demands) raised by politicians and other actors, including claim and actor spans, relations, and polarities. In addition to the annotation GUI, the tool supports the identification of relevant documents, text pre-processing, user management, integration of external knowledge bases, annotation comparison and merging, statistical analysis, and the incorporation of machine learning models as “pseudo-annotators”.

To extract such information from text, we can often utilize the main NLP tools introduced in Section~\ref{sec:nlp_tools}, including text classification, topic modeling, event extraction and score prediction (especially text scaling to predict left-to-right ideology). In NLP literature, social media, such as Twitter, is a popular source of textual data to collect public opinions  \citep{thelwall2011sentiment,paltoglou2012twitter,pak-paroubek-2010-twitter,arunachalam-sarkar-2013-new,rosenthal2015semeval}.

\myparagraph{Political Usage.}
Such information extracted from data is highly valuable for political usage. For example, voters' sentiment, stance, and ideology are important supplementary for traditional polls and surveys to gather information about the constituents' political leaning. Identifying the needs expressed by people is another important survey target, which helps politicians understand what needs they should take care of, and match the needs and availabilities of resources \citep{hiware-etal-2020-narmada}.

Among more specific political uses is to understand the public opinion on parties/president, as well as on certain topics. The public sentiment towards parties \citep{pla-hurtado-2014-political} and President \citep{marchetti-bowick-chambers-2012-learning} can serve as a supplementary for the traditional approval rating survey, and stances towards certain topics \citep{gottipati-etal-2013-learning,stefanov-etal-2020-predicting,luo-etal-2020-detecting} can be important information for legislators to make decisions on debatable issues such as abortion, taxes, and legalization of same-sex marriage.
Many existing studies use NLP on social media text to predict election results \citep{oconnor2010tweets,beverungen2011evaluating,unankard2014predicting,mohammad2015sentiment,tjong-kim-sang-bos-2012-predicting}.
In general, big-data-driven analyses can facilitate decision-makers to collect more feedback from people and society, enabling policymakers to be closer to citizens, and increase transparency and engagement in political issues \citep{arunachalam-sarkar-2013-new}.

\subsection{Interpreting Political Decisions}\label{sec:use2}

\begin{figure}
    \centering
    \includegraphics[width=\textwidth]{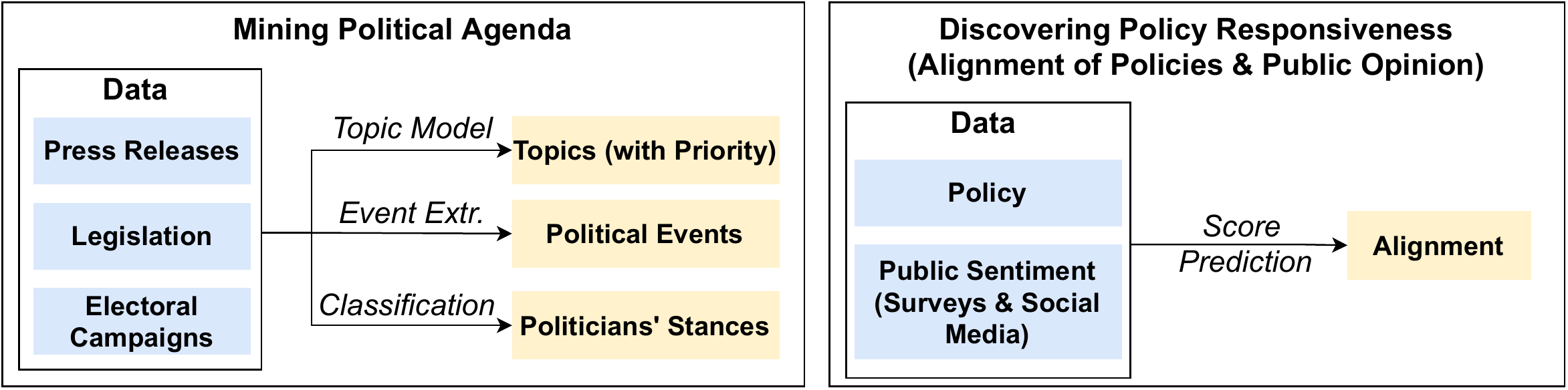}
    \caption{NLP to interpret political decisions.}
    \label{fig:use_interpret}
\end{figure}

After policies are made, political scientists and social scientists can use textual data to interpret political decisions. 
As in Figure~\ref{fig:use_interpret}, there are two major use cases: mining political agendas, and discovering policy responsiveness.

\myparagraph{Mining Political Agendas.}
Researchers can use textual data to infer a political agenda, including the topics that politicians prioritize, political events, and different political actors' stances on certain topics.
Such data can come from press releases, legislation, and electoral campaigns. 
Example of previous studies to analyze the topics and prioritization of political bodies include the research on the prioritization each Senator assigns to topics using press releases \citep{grimmer2010representational}, topics in different parties' electoral manifestos \citep{glavas-etal-2017-cross}, topics in EU parliament speeches \citep{lauscher2016entities} and other various types of text \citep{king2003automated,hopkins2010method,grimmer2010bayesian,roberts2014structural}, as well as political event detection from congressional text and news \citep{nanni2017building}.

Research on politicians' stances include identifying policy positions of politicians \citep[\textit{inter alia}]{winter1977content,laver2003extracting,slapin2008scaling,lowe2011scaling},
how different politicians agree or disagree on certain topics in electoral campaigns \citep{menini-tonelli-2016-agreement}, 
and assessment of political personalities \citep{immelman1993assessment}.

% topics in the press releases can be used to measure the attention each senator allocates to the priorities \citep{grimmer2010representational}.

% Detecting political events: \citep{nanni2017building} from New York Times Corpus and the US Congressional Record, and we
% test its performance on the TREC KBA Stream corpus, and prominent topics in political texts \citep{lauscher2016entities}

% Topics discussed in debates: cross-lingual topical coding of sentences from electoral manifestos of political parties in different languages \citep{glavas-etal-2017-cross}. automated comparison of points of view between two politicians (detecting agreement and disagreement) during an electoral campaign \citep{menini-tonelli-2016-agreement}

Further studies look into how political interests affect legislative behavior. Legislators tend to show strong personal interest in the issues that come before their committees \citep{fenno1973congressmen}, and \citet{mayhew2004congress} identifies that Senators replying on appropriations secured for their state have a strong incentive to support legislations that allow them to secure particularistic goods.

% political interest and legislative behavior: \citep{mayhew2004congress} (e.g., ``Senators who rely upon appropriations secured for their state in press releases have strong incentive to support institutions that allow them to continue to secure particularistic goods''). and legislators are likely to have strong personal interest in the issues that come before committees they lead \citep{fenno1973congressmen}''

\myparagraph{Discovering Policy Responsiveness.}
Policy responsiveness is the study of how policies respond to different factors, such as how changes in public opinion lead to responses in public policy \citep{stimson1995dynamic}. One major direction is that politicians tend to make policies that align with the expectations of their constituents, in order to run for successful re-election in the next term \citep{canes2002out}.
Studies show that policy preferences of the state public can be a predictor of future state policies \citep{caughey2018policy}.
For example, \citet{lax2009gay} show that more LGBT tolerance leads to more pro-gay legislation in response.
% Most policies and public opinion studied in existing literature are often long-term and gradual, taking several decades to observe \citep{lax2009gay,lax2012democratic,caughey2018policy}. The main reasons are that adapting a new policy might take more than a decade \citep{lax2012democratic}.
% due to partisan considerations and institutional settings such as filibuster and veto power delaying the legislative agenda \citep{cox2005setting,wawro2006filibuster,krehbiel2010pivotal}, and observing shifts in public opinion on salient issues also take time \citep{lax2012democratic}.

% Policy responsiveness is the study of the factors that policies respond to \citep{stimson1995dynamic}. One major direction is that politicians tend to make policies that align with the expectations of their constituents, in order to run successful re-election in the next term \citep{canes2002out}.

% Prior research on political decision making addresses the relationship between policy and public sentiments. ``If public opinion changes and then public policy responds, this is dynamic representation (policy responsiveness)'' \citep{stimson1995dynamic}. Motivated by re-election, politicians make policies that align with the expectations of their constituents. Those who went out of step of constituents' preferences will be punished in the next election \citep{canes2002out}.

A recent study by \citet{jin-etal-2021-mining-cause} uses NLP to analyze over 10 million COVID-19-related tweets targeted at US governors; using classification models to obtain the public sentiment, they study how public sentiment leads to political decisions of COVID-19 policies made by US governors. Such use of NLP on massive textual data contrasts with the traditional studies of policy responsiveness which span over several decades and use manually collected survey results \citep{caughey2018policy,lax2009gay,lax2012democratic}. 

% \textit{long-term} setting, where the policies are collected over a span of several decades, e.g., \citet{caughey2018policy}'s collection of public opinion surveys and state policymaking data over 1936-2014, and \citet{lax2009gay}'s collection of public opinion polls and gradual policy changes over 1999-2008. Second, the data sources of existing studies are mostly surveys and polls, which can be time-consuming and expensive to collect \citep{lax2012democratic}. Third, the resulting data are often of relatively small sizes, for both the number of policies and the number of public opinion. 

\subsection{Improving Policy Communication with the Public}\label{sec:use3}

\begin{figure}[t]
    \centering
    \includegraphics[width=\textwidth]{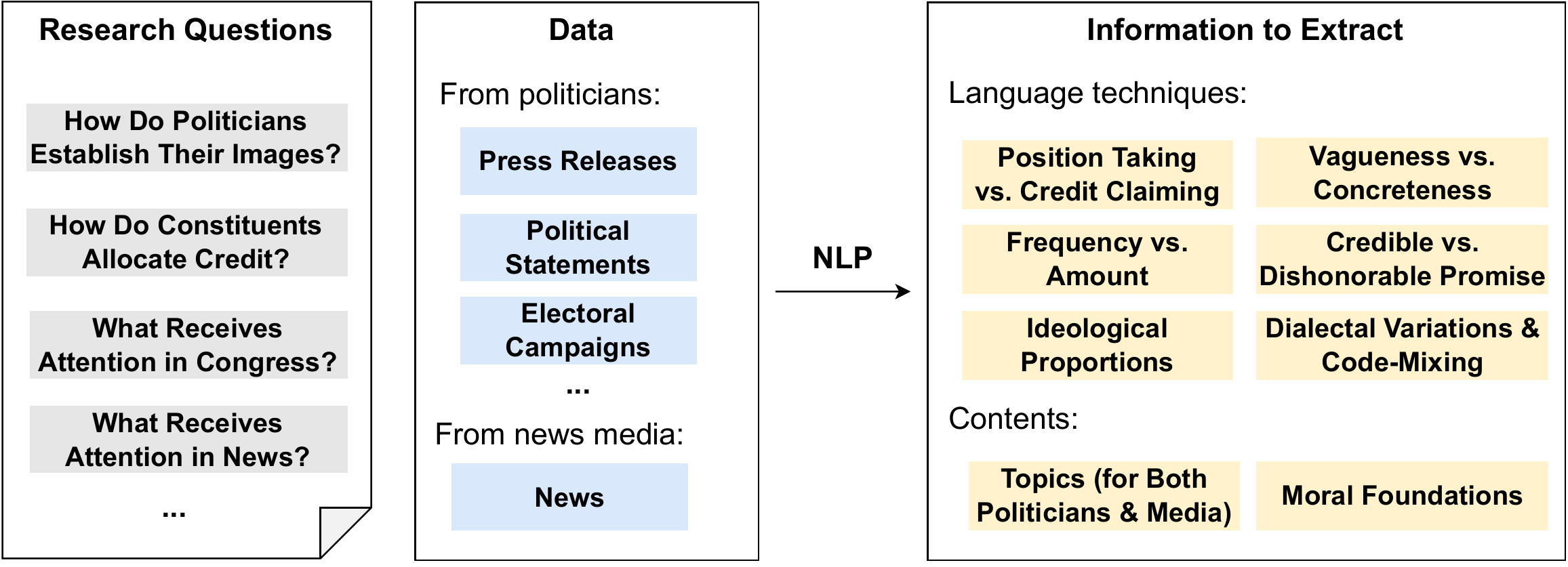}
    \caption{NLP to analyze policy communication.}
    \label{fig:use_comm}
\end{figure}

Policy communication is the study to understand how politicians present the policies to their constituents. As in Figure~\ref{fig:use_comm}, common research questions in policy communication include how politicians establish their images \citep{fenno1978home} such as campaign strategies \citep{petrocik1996issue,simon2002winning,sigelman2004avoidance}, how constituents allocate credit, what receives attention in Congress \citep{sulkin2005issue}, and what receives attention in news articles \citep{semetko2000framing,mccombs2004setting,armstrong2006whose}.

Based on data from press releases, political statements, electoral campaigns and news articles,\footnote{Other data sources used in policy communication research include surveys of Senate staffers \citep{cook1988press}, newsletters that legislators send to constituents \citep{lipinski2009congressional} and so on.} researchers usually analyze two types of information: the language techniques politicians use, and the contents such as topics and underlying moral foundations in these textual documents. 

\myparagraph{Language Techniques.}
Policy communication largely focuses on the types of languages that politicians use. Researchers are interested in first analyzing the language techniques in political texts, and then, based on these techniques, researchers can dive into the questions of why politicians use them, and what are the effects of such usage. 

For example, previous studies analyze what portions of political texts are position-taking versus credit-claiming \citep{grimmer2012words,grimmer2013appropriators}, whether the claims are vague or concrete \citep{baerg2018central,eichorst2019resist}, the frequency of credit-claiming messages versus the actual amount of contributions
% affect constituents' credit assignment more
\citep{grimmer2012words}, and whether politicians tend to make credible or dishonorable promises \citep{grimmer2010representational}. Within the political statements, it is also interesting to check the ideological proportions \citep{sim-etal-2013-measuring}, and how politicians make use of dialectal variations and code-mixing \citep{sravani-etal-2021-political}.

The representation styles usually affect the effectiveness of policy communication, such as 
the role of language
ambiguity in framing the political agenda \citep{page1976theory,campbell1983ambiguity}, and the effect of credit-claiming messages on constituents' allocation of credit \citep{grimmer2012words}.

% how politicians balance position taking and credit claiming affects the effectiveness of the representation,
% and credit claiming messages affect constituents' allocation of credit and the constituents are more sensitive to the total number of messages than the actual amount claimed \citep{grimmer2012words}.

\myparagraph{Contents.}
The contents of policy communication include the topics in the political statements, such as what Senators discuss in floor statements \citep{hill2002symbolic}, and what Presidents address in daily speeches \citep{lee2008dividers}, and also the moral foundations used by politicians underlying their political tweets \citep{johnson-goldwasser-2018-classification}.

Using the extracted content information, researchers can explore further questions such as whether competing politicians or political elites emphasize the same issues  \citep{petrocik1996issue,gabel2007estimating}, and how the priorities politicians articulate co-vary with the issues discussed in
the media \citep{bartels1996politicians}. Another open research direction is to analyze the interaction between newspapers and politicians' messages, such as how often newspapers cover a certain politician's message and in what way, and how such coverage affects incumbency advantage.

% Advanced topics: Measuring
% how often newspapers cover elite statements would provide an answer to a number of
% theoretically important questions, including how reliant local newspapers are on information from Senate offices, identifying the role of partisanship in determining how often a newspaper prints a legislator’s message, and determining how a newspaper’s
% reliance upon information from Congressional offices influences the incumbency
% advantage.

\myparagraph{Meaningful Future Work.}
Apart from analyzing the language of existing political texts that aims to maximize political interests, an advanced question that is more meaningful to society is how to improve policy communication to steer towards a more beneficial future for society as a whole. There is relatively little research on this, and we welcome future work on this meaningful topic.

\subsection{Investigating Policy Effects}\label{sec:use4}

After policies are taken into effect, it is important to collect feedback or evaluate the effectiveness of policies. Existing studies evaluate the effects of policies along different dimensions: one dimension is the change in public sentiment, which can be analyzed by comparing the sentiment classification results before and after policies, following a similar paradigm in Section~\ref{sec:use1}.
There are also studies on how policies affect the crowd's perception of the democratic process \citep{miller1990voters}.

Another dimension is how policies result in economic changes. \citet{calvo2018winners} investigate the negative consequences of policy volatility that harm long-term economic growth. Specifically, to measure policy volatility, they first obtain main topics by topic modeling on presidential speeches, and then analyze how the significance of topics changes over time.

% \citep{calvo2018winners}The World Bank’s Poverty and Equity Global Practice Group used LDA topic modeling to measure changes in policy priorities by examining presidential speeches in 10 Latin American countries and Spain from 1819 to 2016. Using LDA, the authors could identify the main topics for each document and indicate the variation in their significance across countries and over time. In Peru, for instance, topics on infrastructure and public services diminished in importance over time. With the help of topic modeling, the authors were able to establish, for each nation, a negative correlation between policy volatility and long-term growth.
% Although there is wide recognition of the negative consequences of policy volatility for countries’ long-term economic
% growth, there is limited empirical work on this subject. One
% of the reasons is the difficulty of measuring policy volatility
% over long periods of time, especially in developing countries.
% This paper contributes to this literature by constructing
% a proxy for policy volatility that exploits the information
% content of the priorities conveyed in presidential speeches.
% The study creates a policy volatility measure using a Latent
% Dirichlet Allocation algorithm on a novel data set of 953
% presidential speeches in 10 Latin American countries
% and Spain. The paper shows that the proxy for policy
% volatility is negatively correlated with long-term growth
% over 1940–2010. The results are robust to a large set of
% changes in the construction of the proxy for policy volatility.

\section{Limitations and Ethical Considerations} \label{sec:ethics}

There are several limitations that researchers and policymakers need to take into consideration when using NLP for policymaking, due to the data-driven and black-box nature of modern NLP. First, the effectiveness of the computational models relies on the quality and comprehensiveness of the data. Although many political discourses are public, including data sources such as news, press releases, legislation, and campaigns, when it comes to surveying public opinions, social media might be a biased representation of the whole population. Therefore, when making important policy decisions, the traditional polls and surveys can provide more comprehensive coverage. Note that in the case of traditional polls, NLP can still be helpful in expediting the processing of survey answers. 

The second concern is the black-box nature of modern NLP models. We do not encourage decision-making systems to depend fully on NLP, but suggest that NLP can assist human decision-makers. Hence, all the applications introduced in this chapter use NLP to compile information that is necessary for policymaking instead of directly suggesting a policy. Nonetheless, some of the models are hard to interpret or explain, such as text classification using deep learning models \citep{yin-etal-2019-benchmarking,brown2020language}, which could be vulnerable to adversarial attacks by small paraphrasing of the text input \citep{jin2020bert}. In practical applications, it is important to ensure the trustworthiness of the usage of AI. There could be a preference for transparent machine learning models if they can do the work well (e.g., LDA topic models, and traditional classification methods using dictionaries or linguistic rules), or tasks with well-controlled outputs such as event extraction to select spans of the given text that mention events. In cases where only the deep learning models can provide good performance, there should be more detailed performance analysis (e.g., a study to check the correlation of the model decisions and human judgments), error analysis (e.g., different types of errors, failure modes, and potential bias towards certain groups), and studies about the interpretability of the model (e.g., feature attribution of the model, visualization of the internal states of the model).

Apart from the limitations of the technical methodology, there are also ethical considerations arising from the use of NLP. Among the use cases introduced in this chapter, some applications of NLP are relatively safe as they mainly involve analyzing public political documents and fact-based evidence or effects of policies. However, others could be concerning and vulnerable to misuse. For example, although effective, truthful policy communication is beneficial for society, it might be tempting to overdo policy communication and by all means optimize the votes. As it is highly important for government and politicians to gain positive public perception, overly optimizing policy communication might lead to propaganda, intrusion of data privacy to collect more user preferences, and, in more severe cases, surveillance and violation of human rights.
Hence, there is a strong need for policies to regulate the use of technologies that influence public opinions and pose a challenge to democracy.

\section{Conclusions}

This chapter provided a brief overview of current research directions in NLP that provide support for policymaking. We first introduced four main NLP tasks that are commonly used in text analysis: text classification, topic modeling, event extraction, and text scaling. We then showed how these methods can be used in policymaking for applications such as data collection for evidence-based policymaking, interpretation of political decisions, policy communication, and investigation of policy effects. We also discussed potential limitations and ethical considerations of which researchers and policymakers should be aware.

NLP holds significant promise for enabling data-driven policymaking. In addition to the tasks overviewed in this chapter, we foresee that other NLP applications, such as text summarization (e.g., to condense information from large documents), question answering (e.g., for reasoning about policies), and culturally-adjusted machine translation (e.g., to facilitate international communications), will soon find use in policymaking. The field of NLP is quickly advancing, and close collaborations between NLP experts and public policy experts will be key to the successful use and deployment of NLP tools in public policy. 